\pdfoutput=1

\documentclass[11pt]{article}

\usepackage[final]{acl}
\usepackage{graphicx}
\usepackage{times}
\usepackage{latexsym}

\usepackage[T1]{fontenc}

\usepackage[utf8]{inputenc}

\usepackage{microtype}

\usepackage{inconsolata}

\title{NLU-STR at SemEval-2024 Task 1: Generative-based Augmentation and Encoder-based Scoring for Semantic Textual Relatedness}

\author{Sanad Malaysha, Mustafa Jarrar, Mohammed Khalilia \\
   Birzeit University, Palestine \\
    \texttt{\{smalaysha, mjarrar, mkhalilia\}@birzeit.edu} \\
 }
 
\begin{document}
\maketitle
\begin{abstract}
Semantic textual relatedness is a broader concept of semantic similarity. It measures the extent to which two chunks of text convey similar meaning or topics, or share related concepts or contexts. This notion of relatedness can be applied in various applications, such as document clustering and summarizing. SemRel-2024, a shared task in SemEval-2024, aims at reducing the gap in the semantic relatedness task by providing datasets for fourteen languages and dialects including Arabic. This paper reports on our participation in Track A (Algerian and Moroccan dialects) and Track B (Modern Standard Arabic). A BERT-based model is augmented and fine-tuned for regression scoring in supervised track (A), while BERT-based cosine similarity is employed for unsupervised track (B). Our system ranked 1\textsuperscript{st} in SemRel-2024 for MSA with a Spearman correlation score of 0.49. We ranked 5\textsuperscript{th} for Moroccan and 12\textsuperscript{th} for Algerian with scores of 0.83 and 0.53, respectively.
\end{abstract}

\section{Introduction}
\label{sec:intro}
The literature commonly examines semantic similarity, where the focus is on whether two linguistic units (words, phrases, sentences, etc.) share similar meanings \citep{BentivogliBMMBZ16}. However, semantic textual relatedness (STR) is less explored due to its complexity and the scarcity of datasets \citep{AbdallaVM23,DH21}. While the former task checks for the presence of similar meaning or paraphrase, STR takes a more comprehensive approach, evaluating relatedness across multiple dimensions, spanning topical similarity, conceptual overlap, contextual coherence, pragmatic connection, themes, scopes, ideas, stylistic conditions, ontological relations, entailment, temporal relation, as well as semantic similarity itself \citep{miller1991contextual, halliday2014cohesion,J21, J11}. For example, consider the two sentences (\textit{The Earth orbits the sun at a speed of \textasciitilde110,000 km/h.}) and (\textit{Earth rotates at \textasciitilde1670 km/h around its axis.}). They hold semantic relatedness through the shared topic of Earth's speeds. In contrast, both sentences are not semantically similar as they possess distinct meanings. This illustrates the broader range of STR as described by \citet{AbdallaVM23}, which ranges from highly relevant sentences, expressing the same idea with different wording, to entirely unrelated sentences, discussing unrelated topics.

Semantic relatedness has proven to be useful in evaluating sentence representations generated by language models \citep{AsaadiMK19}, in addition to question answering \citep{TsatsaronisVV14}, machine translation \citep{MiX24}, plagiarism detection \citep{SabirMP19}, word-sense disambiguation \citep{arabglossbert, MalayshaJK23}, among others. Exploring the relatedness and similar tasks in languages other than English is hindered by the lack of data \citep{JMHK23,HJ21}. The SemRel-2024 shared task \citep{semrel2024dataset} provided datasets in fourteen languages and offered three tracks. In the supervised track (A), training and testing are performed on the same language. In the unsupervised track (B), the use of labeled data for training is prohibited; and in the cross-lingual track (C), testing is conducted on a different language than the one used for training.

This paper presents our contribution to track A and track B. 
In track A, we fine-tuned BERT models using the Algerian and Moroccan sentence pairs to produce similarity scores. To enrich the data, we augmented the SemRel-2024 dataset \citep{semrel2024dataset} by generating additional sentence pairs from Google Gemini \footnote{https://gemini.google.com/}, a generative model, using a predefined prompt template. These generated pairs imitated the style and meaning of the existing pairs, and we assigned them scores corresponding to the originals. We used the same datasets provided by the Shared Task in addition to a \textasciitilde760 augmented Moroccan pairs to fine-tune BERT models, AraBERTv2 \citep{antoun2020arabert} and ArBERTv2 \citep{Abdul-MageedEN20}, which resulted in a performance enhancement of 0.05 points. 
In track B, as training on labeled data is not allowed, we used cosine similarity using average pooling embedding \citep{ZhaoZHCY22} on top of each model. 
Our approaches achieved Spearman scores \citep{TsatsaronisVV14} of 0.49 for MSA (ranked first), 0.83 for Moroccan (ranked fifth), and 0.53 for Algerian (ranked twelfth).

\section{Related Work}
\label{sec:related-work}


Semantic textual relatedness (STR) has proven to be a valuable task in numerous NLP applications, including the evaluation of LLMs \citep{AsaadiMK19, NaseemRKP21}. Determining the degree of relatedness in STR, however, remains a challenging task in computational semantics. That is because STR encompasses a broader range of commonalities beyond just meaning, including shared viewpoint, topic, and period, demanding a deeper understanding than semantic similarity alone \citep{AsaadiMK19, AbdallaVM23}. For example, consider reading these two sentences (\textit{He heard the waves crashing gently}) and (\textit{Making him feel calm and peaceful}). While humans easily recognize their strong relatedness and shared description of the same view (a beach scene), machines require advanced lexical and statistical methods to achieve the same level of understanding. STR techniques mainly come from four approaches: lexical similarity \citep{ChenSY18a, jarrardb, ADJ19}, semantic similarity \citep{hasan2020knowledge,GJJB23}, deep learning \citep{zhang2019multi}, and LLMs \cite{LiYC21}.

Recently, \citet{AbdallaVM23} introduced their STR-2022 dataset, which uses fine-grained scores ranging from 0 (least related) to 1 (completely related). Their dataset consists of 5,500 scored English sentence pairs. They framed the task as supervised regression, where they fine-tuned two language models, BERT-base \citep{kenton2019bert} and RoBERTa-base \citep{roberta}, and applied average pooling on top of the final embedding layer. Their testing of these models on the STR-2022 dataset yielded an average Spearman correlation of 0.82 for BERT-base and 0.83 for RoBERTa-base. On the other hand, their unsupervised experiments using Word2Vec \citep{word2vec} achieved a correlation score of 0.60, outperforming both BERT-base (0.58) and RoBERTa-base (0.48) by 0.02 and 0.12 points, respectively.

\citet{AsaadiMK19} created the Bi-gram Semantic Relatedness Dataset (BiRD) for examining semantic composition. To avoid inconsistencies and biases from traditional 1-5 rating scales, they employed fine-grained scoring of bi-gram pairs (0-1) using the best-worst scaling (BWS) annotation technique \citep{KiritchenkoM17}. The dataset consists of 3,345 scored English term pairs. They utilised three models to generate word representations: GloVe \citep{glove}, FastText \citep{fasttext}, and a word-context co-occurrence matrix \citep{TurneyNAC11}. To calculate relatedness scores between pairs, they employed cosine similarity between the generated addition-pooled vectors. The FastText model achieved the highest performance with a Pearson correlation of 0.60.


The semantic relatedness between noun-pairs was studied using contextual similarity by \citet{miller1991contextual}. They attempted to understand distinctions between nouns in contextual discourse and how the similarity can be broader than just the meaning. Additional ideas could rely on extracting named entities \citep{LJKOA23,JKG22} to measure the relatedness \cite{GhoshDKZA23}. However, the task evolved, leading to the creation of the up-to-date dataset presented by the SemRel-2024 shared task \citep{semrel2024task}. Their dataset annotation scores are at the level of sentence pairs. They shared baseline results for fourteen languages and dialects using Spearman correlation score. Since our focus is on Arabic, we have chosen its results to show. For example, their baseline is 0.42 for MSA in track B using multi-lingual BERT (mBERT) \citep{kenton2019bert}, 0.60 for Algerian and 0.77 for Moroccan in track A using Label Agnostic BERT Sentence embeddings (LaBSE) \citep{FengYCA022}. Specifically, their Algerian Arabic dataset offers 1,261 training and 583 test instances, Moroccan Arabic dataset includes 924 training and 425 test instances, and MSA Arabic dataset has 595 instances for testing.

Many efforts have been made to understand Arabic dialects, such as dialect identification, intent detection, and morphological annotations \cite{EJHZ22, ANMFTM23, JZHNW23, JHRAZ17, JBKEG23}, but none studied STR between dialects.


\begin{figure}[ht!]
    \centering
    \includegraphics[scale=0.40]{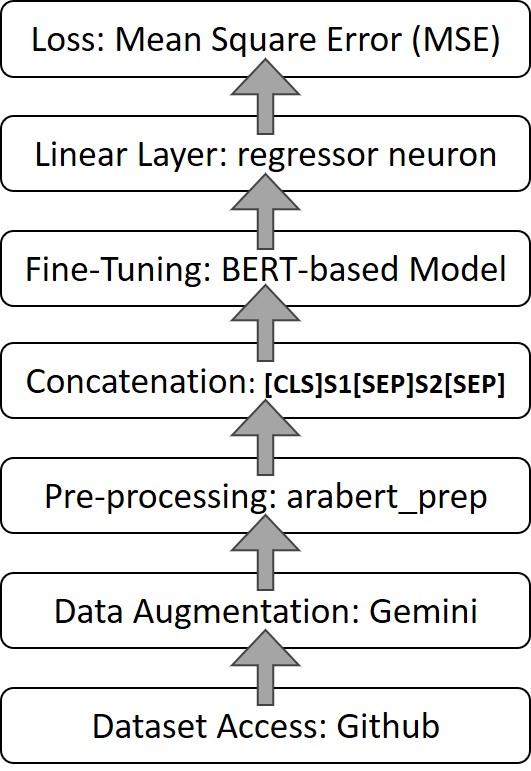}
    \caption{BERT-based Supervised Architecture (A).}
    \label{fig:supervised-arch}
\end{figure}

\section{System Overview}
\label{sec:system-overview}
This section presents the techniques, datasets, and the augmentation we employed in tracks A and B.

\subsection{Supervised Track (A)}
\label{sub:supervised}

Since the datasets use continuous scoring values, we tackled STR as a regression problem. We fine-tuned BERT with Mean Squared Error (MSE) objective. The model uses a regressor output layer, represented by a single neuron to predict the scores of the sentence-pairs. The data was pre-processed using the technique presented in \citep{antoun2020arabert} to achieve standardized word forms. Before supplying the sentence pairs to the model, each was concatenated using the special tokens of the model input in this format:\texttt{[CLS]Sentence1[SEP]Sentence2[SEP]}. Figure \ref{fig:supervised-arch} depicts our method architecture for the supervised track (A). Since we focused on the Algerian and Moroccan dialects in this track, we investigated various model parameters including learning rates, number of epochs, and pre-trained models to understand which model is better suited for each dialect. We found that both models, AraBERTv2 \footnote{https://github.com/aub-mind/arabert} and ArBERTv2 \footnote{https://huggingface.co/UBC-NLP/ARBERTv2}, best fits the Moroccan dialect more than Algerian. Nonetheless, we used same models for the Algerian dataset.

\begin{figure}[ht!]
    \centering
    \includegraphics[scale=0.40]{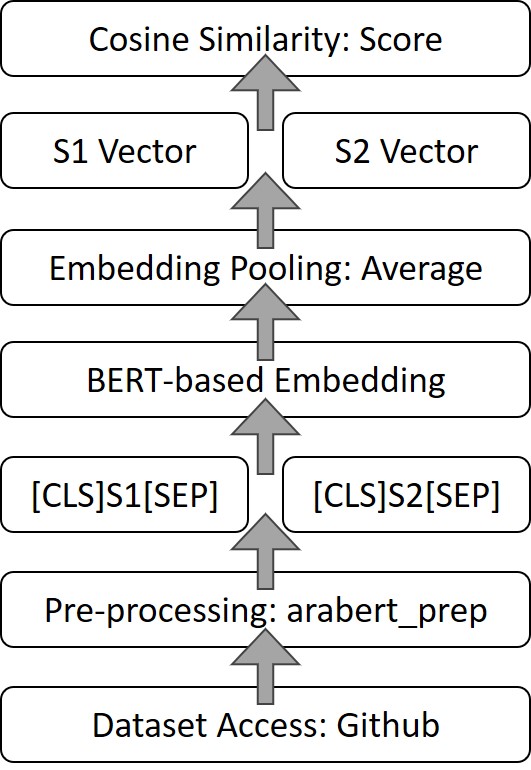}
    \caption{BERT-based Unsupervised Architecture (B).}
    \label{fig:unsupervised-arch}
\end{figure}

\begin{figure*}[ht!]
    \centering
    \includegraphics[scale=0.650]{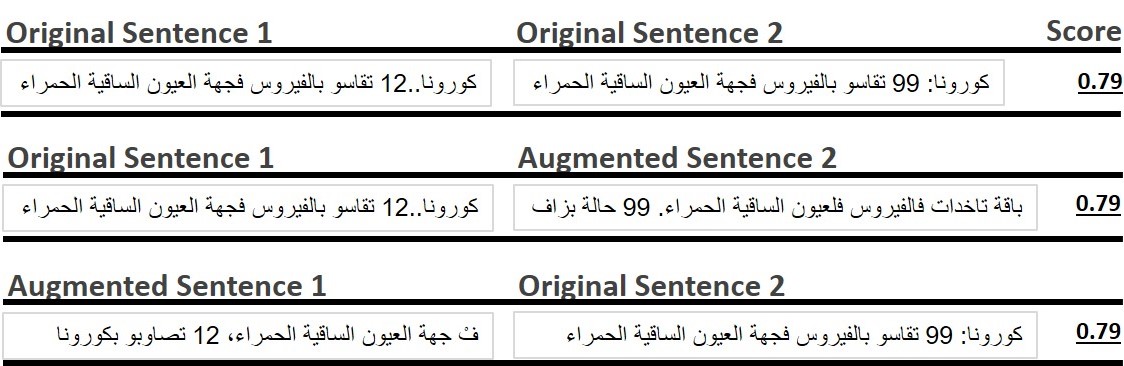}
    \caption{Example of the augmented sentence-pairs.}
    \label{fig:augments}
\end{figure*}

\subsection{Unsupervised Track (B)}
\label{sub:unsupervised}
The STR using MSA is covered in track  B (unsupervised learning), where training (or fine-tuning) on labeled data is not permitted. We employed cosine similarity \citep{cosinesimilarity} as an unsupervised technique to calculate the sentence-pair scores. Figure \ref{fig:unsupervised-arch} illustrates our architecture. We conducted initial experiments using the same aforementioned models, ArBERTv2 and AraBERTv2, for generating sentence representations. Various pooling options (CLS, average, max, and min) \citep{ZhaoZHCY22} were applied on the final embedding layer in each (frozen) model, and found that AraBERTv2 with average-pooling is better suited for MSA in this track. The same data pre-processing used in track A is applied in B.

\subsection{Datasets}
\label{sub:data-augmentation}

The datasets provided by the SemRel-2024 shared task cover fourteen languages and dialects. In the paper, we used three Arabic datasets (Algerian, Moroccan, and MSA). Table \ref{table:datasets} presents their data splits, including train, development, and testing. MSA has no labeled train data as it is included in Track B. However, for the other two dialects, we employed BERT-based models, that requires large train data \citep{bevilacqua2021recent}. 

\begin{table}[ht!]
\centering
\begin{tabular}{llll}
\hline
               & \textbf{MSA} & \textbf{Algerian} & \textbf{Moroccan} \\ \hline
\textbf{Train} &              &                   &                   \\
\hspace{5pt}Original       & --           & 1,261             & 924               \\
\hspace{5pt}Augments       & --           & --                & 757               \\
\hspace{5pt}Total          & --           & 1,261             & 1,681             \\ \hline
\textbf{Dev.}   & 32           & 97                & 70                \\ \hline
\textbf{Test}  & 595          & 583               & 425               \\ \hline
\end{tabular}
\caption{\label{table:datasets}
The original and augmented datasets splits.
}
\end{table}

\begin{figure}[ht!]
    \centering
    \includegraphics[scale=0.6]{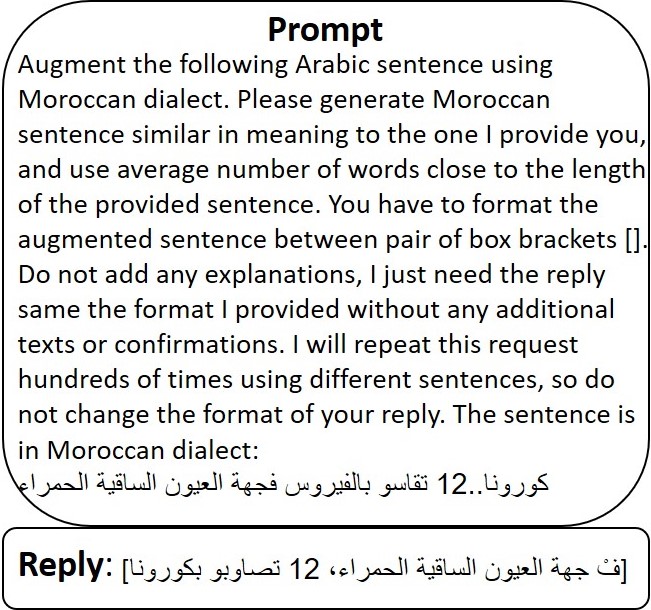}
    \caption{The prompt template employed for Gemini.}
    \label{fig:prompt}
\end{figure}

Different methods can be used for data augmentation, such as back-translation \cite{lin2021context} and generative models \citep{SaidiJKS22}. The back-translation technique was tested by \citep{MalayshaJK23} and showed minor improvement in performance. The availability of high-quality generative models, such as ChatGPT \footnote{https://chat.openai.com/} and Google Gemini, encouraged us to employ them in automatic augmentation. We employed in-context learning \citep{incontextlearning} by prompting both models with the request depicted in Figure \ref{fig:prompt}.


The initial manual reviews and tests for twenty prompts of Moroccan and Algerian sentences showed that both models are weak in Algerian comprehension. ChatGPT is also weak in the Moroccan, while Gemini demonstrated a high understanding of the Moroccan. Therefore, we decided to employ Gemini to augment the Moroccan train split. From every sentence-pair, we took each sentence and prompted it using the template in Figure \ref{fig:prompt}. We mapped the augmented (new) sentence from the model with the other sentence in the same pair using the same score of the pair, as illustrated in Figure \ref{fig:augments}. By manually reviewing all the model replies, we found cases that were not valid (wrong content), and accordingly, we defined filters to exclude the not applicable data per the following rules:
\begin{itemize}
    \item The model admits in the reply that it is just a language model and cannot fulfill the request. The model reply in such case has common format to rely on for the filter comparison.
    \item The case when the reply goes far from the original meaning. This option is achieved by manually reviewing the paraphrased contents.
    \item When the model rejects augmentation because the requested sentence contains information that breaks the model policy, i.e., talking about public figures or sensitive discussions. Similar to first rule, it has common reply format to automatically compare with.
\end{itemize}
Finally, after filtering the invalid augmentations, we reached 757 accepted sentences which we added to the Moroccan training set (See Table \ref{table:datasets}), reaching a total of 1,681 instances.

\begin{table*}[ht!]
\centering
\begin{tabular}{llll|l}
                   \textit{Development Phase} & \multicolumn{3}{c|}{\textbf{Track A}} & \textbf{Track B} \\ \hline
                   & Algerian            & Moroccan  & Augmented Moroccan      & MSA              \\ \cline{2-5}        \hline
\textbf{ArBERTv2}  & 0.55           & 0.82 & 0.88$\uparrow$ & 0.42             \\ \hline
\textbf{AraBERTv2} & 0.69  & 0.84 & 0.79$\downarrow$          & 0.58    \\ \hline
\end{tabular}
\caption{\label{table:dev-results}
Our results on the development phase (i.e., on development split).
}
\end{table*}

\begin{table*}[ht!]
\centering
\begin{tabular}{llll|l}
                   \textit{TEST Phase} & \multicolumn{3}{c|}{\textbf{Track A}} & \textbf{Track B} \\ \hline
                   & Algerian            & Moroccan  & Augmented Moroccan      & MSA              \\ \cline{2-5} 
\textbf{Baseline} \scriptsize\citep{semrel2024dataset}   & 0.60           & 0.77 & 0.77          & 0.42             \\ \hline
\textbf{ArBERTv2}  & 0.42$\downarrow$           & 0.78$\uparrow$ & \textbf{0.83}$\uparrow$ & 0.34$\downarrow$             \\ \hline
\textbf{AraBERTv2} & \textbf{0.53}$\downarrow$  & 0.79$\uparrow$ & 0.77$\uparrow$          & \textbf{0.49}$\uparrow$    \\ \hline
\end{tabular}
\caption{\label{table:results}
The evaluation results on the test data. Our official ranked scores are in bold.
}
\end{table*}

\section{Experimental Setup}
\label{sec:experimental-setup}
Our experiments fine-tuned two language models for Algerian and Moroccan, where we used the following pre-trained models: maubmindlab/bert-base-arabertv02 \citep{antoun2020arabert} and UBC-NLP/ARBERTv2 \citep{Abdul-MageedEN20}. We employed the training data provided by the shared task, in addition to the data generated by our augmentation technique, when applied. The development data is excluded from either training or testing in the official evaluation phase, and testing is done on the shared task test set (See Table \ref{table:datasets}). The data pairs were concatenated using special tokens (\texttt{[CLS]} and \texttt{[SEP]}), as depicted in Figure \ref{fig:supervised-arch}, and digested by the models. The fine-tuning was done as a regression task using one neuron in the output layer, optimized using MSE as the loss function, and we used R-squared \citep{miles2005r} to measure the improvement. The final hyper-parameters in the fine-tuning process were: 10 epochs for training, 4 epochs for early stopping, a batch size of 16, 512 is the maximum sequence length, a learning rate of $2e^{-5}$, 50 evaluation steps, a seed of 42, and train ($\pm$ augmented data) split.

In the experiments of B track for the MSA, no supervised fine-tuning is needed. Therefore, we neither used labeled data nor augmentation. We employed average-pooling on the embeddings of the sentence tokens from the final layer in each model. Then, we calculated the cosine similarity between the average embeddings of the sentences in each pair. This was done to estimate the fine-grained scores for the test (or development) data provided by the shared task. The shared task considers Spearman correlation score to evaluate the submitted predictions against their ground truth.

\section{Results}
\label{sec:result}
Our approaches have achieved competitive ranks in the SemRel-2024 shared task. The official results of the tracks we participated in, as well as the baselines that were introduced by \citet{semrel2024dataset}, are shown in Table \ref{table:results}. Additionally, our results on the development data are presented in Table \ref{table:dev-results}. In the test evaluation, we ranked first in Track B for the MSA, with a Spearman correlation score of 0.49 using the AraBERTv2 model, outperforming the baseline by 0.07 points. However, ArBERTv2 did not perform well in Track B for MSA on both test and development splits. In contrast, ArBERTv2 achieved a high score in Track A for the Moroccan dialect when fine-tuned on both the train split and augmentation data, outperforming the baseline by 0.06 points on test split, ranking 5th among the submitted systems. Nonetheless, neither of the models, ArBERTv2 or AraBERTv2, surpassed the baseline for the Algerian dialect in Track A, where our rank is 12. Similarly, both models achieved low performance on the Algerian development split. It is possible that if we were able to augment the Algerian data as well, it could have performed better, similar to the improvement achieved in the Moroccan dataset. It is worth noting that AraBERTv2 outperformed both the baseline and ArBERTv2 on the original training data of the Moroccan dataset. However, its performance degraded on both test and development splits once the augmentation was included in the fine-tuning, unlike what happened with the ArBERTv2 model, on both splits. This could be due to the nature of the data utilized in the pre-training phase of the model. Due to the anisotropy problem \citep{BaggettoF22} inherent in BERT-based pre-trained models, we noted that computing cosine similarity directly between sentence representations is insufficient for discerning relatedness.

\section{Conclusion}
\label{sec:conclusion}
We presented our contributions to the SemRel-2024 shared task. We targeted three Arabic dialects covered by the shared task datasets, including MSA, Algerian, and Moroccan. Our approaches employed supervised and unsupervised techniques using commonly known language models, namely ArBERT and AraBERT. We augmented the training data using generative models, which enhanced the models' performance. Our system ranked first (MSA), fifth (Moroccan), and twelfth (Algerian) across the different tracks. We plan to augment additional data of Moroccan and Algerian using other models than what we used in this work. We will use the augmentations to experiment with both Arabic mono-dialect and cross-dialect fine-tuning.




\bibliography{acl_latex}

\end{document}